 \documentclass[pmlr,twocolumn,10pt]{jmlr} % W&CP article

\usepackage{booktabs}
\usepackage{multirow}
\usepackage{siunitx}

\newcommand{\equal}[1]{{\hypersetup{linkcolor=black}\thanks{#1}}}

\jmlrvolume{225}
\jmlryear{2023}
\jmlrsubmitted{225}
\jmlrpublished{225}
\jmlrworkshop{Machine Learning for Health (ML4H) 2023} % W&CP title

 \title[NoteContrast]{NoteContrast: Contrastive Language-Diagnostic Pretraining for Medical Text}

 \author{%
  \Name{Prajwal Kailas}\equal{These authors contributed equally} \Email{pkailas@bwh.harvard.edu}\\
  \addr Brigham and Women's Hospital \\
  \Name{Max Homilius}\footnotemark[1] \Email{mhomilius@bwh.harvard.edu}\\
  \addr Brigham and Women's Hospital, Harvard Medical School \\
  \Name{Rahul C. Deo} \Email{rdeo@bwh.harvard.edu}\\
  \addr Brigham and Women's Hospital, Harvard Medical School \\
  \Name{Calum A. MacRae} \Email{cmacrae@bwh.harvard.edu}\\
  \addr Brigham and Women's Hospital, Harvard Medical School \\
 }

\begin{document}

\maketitle

\begin{abstract}
Accurate diagnostic coding of medical notes is crucial for enhancing patient care, medical research, and error-free billing in healthcare organizations. Manual coding is a time-consuming task for providers, and diagnostic codes often exhibit low sensitivity and specificity, whereas the free text in medical notes can be a more precise description of a patient's status. Thus, accurate automated diagnostic coding of medical notes has become critical for a learning healthcare system. Recent developments in long-document transformer architectures have enabled attention-based deep-learning models to adjudicate medical notes. In addition, contrastive loss functions have been used to jointly pre-train large language and image models with noisy labels. To further improve the automated adjudication of medical notes, we developed an approach based on i) models for ICD-10 diagnostic code sequences using a large real-world data set, ii) large language models for medical notes, and iii) contrastive pre-training to build an integrated model of both ICD-10 diagnostic codes and corresponding medical text. We demonstrate that a contrastive approach for pre-training improves performance over prior state-of-the-art models for the MIMIC-III-50, MIMIC-III-rare50, and MIMIC-III-full diagnostic coding tasks.
\end{abstract}

% \section{Instructions}
% \label{sec:instructions}

% This is the template for the \textbf{Proceedings Track} for the Machine Learning for Health (ML4H) symposium 2023.
% Please follow the below instructions:

% \begin{enumerate}
%     \item The submission in the Proceedings Paper Track is limited to 8 pages.
%     \item Please, use the packages automatically loaded (amsmath, amssymb, natbib, graphicx, url, algorithm2e) to manage references, write equations, and include figures and algorithms. The use of different packages could create problems in the generation of the camera-ready version. Please, follow the example provided in this file.
%     \item References must be included in a .bib file.
%     \item Please write your paper in a single .tex file.
%     \item The manuscript, data and code must be anonymized during the review process.
%     \item For writing guidelines please consider the official ML4H call for papers at \url{https://ml4health.github.io/2023/}
% \end{enumerate}

% \newpage

\section{Introduction}
\label{sec:intro}

Accurate and automated diagnostic annotations of medical notes have become increasingly important in healthcare systems for improving the efficiency of care and enabling large-scale real-world data analyses. While diagnostic codes are vital for healthcare providers for tracking disease incidence and billing, manual coding is often cumbersome, time-consuming, and prone to errors. Machine learning methods have been developed to automate the diagnostic coding of medical notes. This is a complex task, as frequently multiple codes are applicable for a single note or single condition, and there are over 60,000 medical codes of varying specificity in the hierarchical International Classification of Diseases (ICD-10) system. While the large number of different codes allows for detailed and specific documentation of medical conditions, some codes are used distinctively by individual clinicians, or in different locations, and many are used infrequently. Often, only the most relevant codes for billing purposes are used per medical encounter. As a result, accurate coding can be seen as a multi-label problem with noisy training labels and a long tail of rarely applied diagnostic codes.

Recent work has used long-document transformers and contrastive pre-training to annotate notes spanning thousands of tokens. However, these approaches typically depend on pre-defined biomedical ontologies like the Unified Medical Language System (UMLS) \citep{Yang:2022aa, Yuan:2022aa}, or ICD-9 and ICD-10 hierarchies \citep{Xie:2019aa, Cao:2020aa} to derive meaningful distances between different diagnoses, and often involve complex preprocessing. We sought to combine a data-driven contextual embedding of diagnostic codes with a straightforward contrastive pre-training objective to improve the automated annotation of medical notes. For this we developed an approach based on i) contextual embedding models for diagnostic codes based on a large real-world data set ii) large language models for medical notes and iii) contrastive pre-training to build an integrated model of both ICD-10 codes and corresponding medical text. We show that this contrastive approach incorporating real-world data significantly improved performance over prior state-of-the-art approaches using static sources of biomedical information in the MIMIC-50, MIMIC-50 rare and MIMIC-full benchmarks.

\section{Related work}

\subsection{Diagnostic Coding}

Automatic diagnostic coding is a multi-label classification task assigning ICD codes to medical notes, typically ranging from several hundred to more than 2000 words per note. In addition, the label space is large and sparse, with over 60,000 codes in the most recent version of ICD-10, and a long tail of rarely diagnosed conditions. Several studies have proposed different methods to address this problem. Complex patterns between text and ICD codes were learned using variations of LSTM networks, dilated convolutions, residual connections, and per-label attention \citep{Mullenbach:2018aa, Li:2020ab, Ji:2020aa, Vu:2020aa}. Label representations were further improved by using graph convolution networks to capture the hierarchical structure of diagnostic codes \citep{Xie:2019aa, Cao:2020aa, Michalopoulos:2022aa}. Shared representations can be further improved by extracting representations from low- and high-frequency codes via self-distillation \citep{Zhou:2021aa}, and UMLS-based code synonyms have been used to provide more comprehensive knowledge than capturing code hierarchies \citep{Yuan:2022aa}. Other studies have proposed various techniques to improve coding accuracy, such as using pre-trained biomedical language models with segment pooling to encode longer texts \citep{Huang:2022aa}, exploiting the discourse structure of notes by utilizing section and reconciled label embeddings \citep{Zhang:2022aa}, and incorporating tree-based features constructed from structured electronic health record (EHR) data such as lab values and medications as additional embedding vectors \citep{Liu:2022aa}. Self-alignment learning with a hierarchical contrastive loss has been used to inject knowledge from biomedical ontologies, and prompt-based fine-tuning has been shown to be a powerful approach for predicting diagnostic codes \citep{Yang:2022aa}.

\subsection{Language Models for Biomedical Text}
There are many pre-trained transformer-based language models for clinical and biomedical tasks, including those trained on masked language modeling (MLM) as the original BERT architecture \citep{Devlin:2019aa} and other pre-training objectives. Notable MLM models in the biomedical domain include SciBERT \citep{Beltagy:2019aa}, BioBERT \citep{Lee:2019aa}, ClinicalBERT \citep{Alsentzer:2019aa}, and BioLM \citep{Lewis:2020aa}, which are trained on the semantic scholar %TODO check citation \citep{Ammar:2018aa}
, scientific abstracts from PubMed and PMC or the MIMIC-III dataset \citep{Johnson:2016aa}. Most of these models work at a sentence level and can handle up to 512 tokens, but for longer document-level tasks, the Clinical Longformer and Clinical BigBird \citep{Li:2022aa} models can handle up to 4096 tokens and are based on Longformer \citep{Beltagy:2020aa} and BigBird \citep{Zaheer:2022aa} architectures, respectively.

\subsection{Medical Code Representations}
Transformer models have also been used to encode temporal sequences of diagnostic codes across multiple hospital visits. These models are pre-trained with a masked language model objective on diagnostic, billing, and procedure code sequences. BEHRT \citep{Li:2020aa} is a BERT model that uses additional position and segment embeddings to distinguish between visits and encode temporal information. It was trained on EHR data from 1.6 million patients and tested on three disease prediction tasks (disease diagnosis on the next visit, within 6 months, and within 12 months). MedBERT \citep{Rasmy:2021aa} is conceptually similar to BEHRT but encodes both ICD-9 and ICD-10 codes and uses an additional serialization embedding to maintain the relative order of diagnostic codes within a visit. Lastly, G-BERT \citep{Shang:2019aa} combines graph neural networks and BERT to encode ICD-9 diagnostic codes and ATC drug classification codes. The model is evaluated on a multi-label medication recommendation task given medication and diagnostic history.

% \begin{figure}[htbp]
%  % Caption and label go in the first argument and the figure contents
%  % go in the second argument
% \floatconts
%   {fig:nodes}
%   {\caption{Example Image}}
%   {\includegraphics[width=0.5\linewidth]{images/nodes}}
% \end{figure}

\subsection{Contrastive Learning}

\begin{figure*}[h!]
\floatconts
  {fig:method}
  {\caption{A) The input to the diagnosis model is a sequence of ICD-10 codes, where positions and token type ids are relative to a “current encounter” (shown in red). B) The text encoder is a language model for long documents pre-trained on medical notes. C) Contrastive training of corresponding ICD-10 code sequences and medical note pairs. D) Pairs of ICD-10 codes are matched to textual descriptions of the codes as a fine-tuning step. E) For prompt-based classification \citep{Yang:2022aa}, the labels are concatenated with the medical note text. The prompt is a textual description of the ICD-9 code, and the outputs at the masked positions are processed to obtain the final multi-label classification output.}}
  {\includegraphics[width=\textwidth]{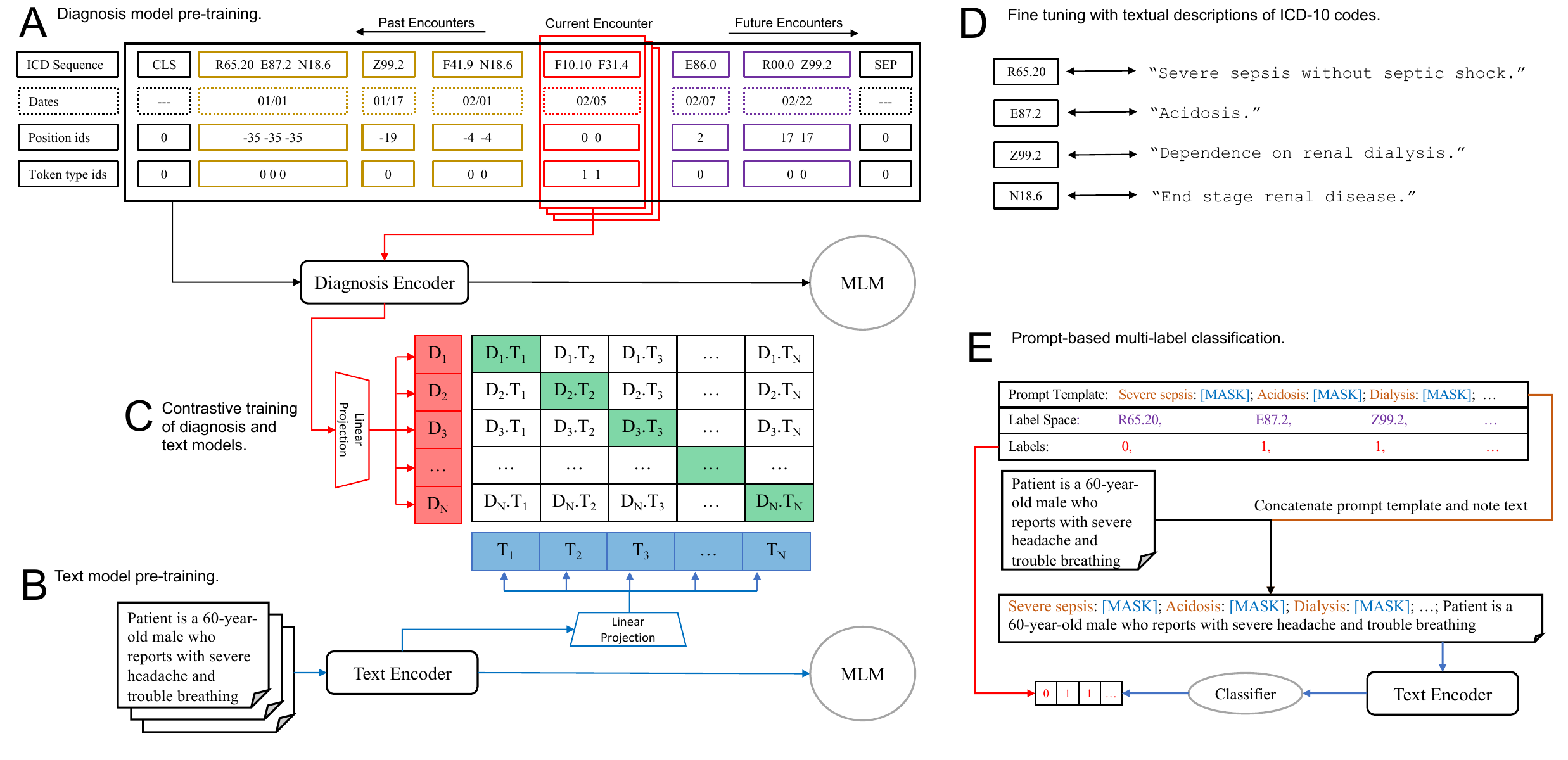}}
\end{figure*}

Contrastive learning involves training a model on a batch of examples that contain positive and negative pairs, maximizing the agreement among positive pairs while minimizing the agreement between negative pairs. This method has been applied for general-purpose pre-training and in specific domains, such as medical imaging. The ConVIRT model \citep{Zhang:2020aa} was an early model to use a contrastive learning objective for medical visual representations, followed by several approaches focusing on medical images with accompanying textual descriptions such as radiology reports \citep{Muller:2021aa, Huang:2021aa, Wang:2021aa, Wang:2022aa}. 
CLIP \citep{Radford:2021aa} is a general large-scale contrastive learning model trained on 400 million image-text pairs from the internet and ALIGN \citep{Jia:2021aa} was trained on over one billion noisy image alt-text pairs. Contrastive approaches have also been applied to the image or text domain separately; for example, SimCLR \citep{Chen:2020aa} is a framework for visual representations based on image-image contrastive loss and augmentations. Text-text contrastive approaches sample pairs based on neighboring text segments on the internet \citep{Neelakantan:2022aa} or independent cropping data augmentations \citep{Izacard:2021aa} and have achieved state of the art results on classification, semantic search, sentence similarity, and retrieval. Contrastive learning is increasingly being applied to specific domains and shows promising results in improving performance on various tasks. For example, SCEHR \citep{Zang:2021aa} is a contrastive learning framework based on EHR time series data applied to clinical risk prediction problems.

\section{Methods}

% REMOVED: subsection heading
%\subsection{Contrastive Training of Medical Text and Diagnostic Code Sequences}

%We propose learning diagnostically relevant representations of medical notes by aligning medical text with corresponding ICD-10 diagnostic code sequences.
We propose learning diagnostically relevant representations of medical notes by aligning medical text with corresponding sequences of one or more ICD-10 diagnostic codes used during the same clinical encounter. 
We use a contrastive learning approach for pre-training the model, where the associated ICD-10 diagnostic codes of a medical note are used as positive signal and contrasted against the diagnostic codes belonging to other medical notes. We describe this approach as a combination of three components, an encoder for sequences of ICD-10 diagnostic codes, an encoder for medical text, and a joint model for contrastive training and alignment of these components, each of which are described in more detail below.

\subsection{Modeling ICD-10 Sequences}

We trained a RoBERTa model \citep{Liu:2019aa} on temporal sequences of diagnostic codes using real-world data of a large patient cohort. To learn long-term temporal associations as well as co-occuring diagnoses, we used sequences of ICD-10 codes across multiple clinical encounters of a patient over time. We selected one encounter as the “encounter of interest”, and calculated the time difference (in days) for past and future encounters, with 0 indicating all diagnostic codes in the current encounter. These relative position values were used to calculate positional embeddings based on sine and cosine functions of different frequencies \citep{Vaswani:2017aa}. In addition, token type identifiers distinguished between the current encounter and others. This approach allowed us to encode past diagnostic history, future events, and patterns in the sequences. % \ref{fig:method}
Figure \ref{fig:method}A shows an example of the sequence of ICD-10 codes for a single patient, who had 6 hospital encounters in their history that have been coded, and the 4th encounter was randomly selected to be the current encounter.
We consider each ICD-10 code as a single token, and trained the ICD-10 sequence model using the masked language modeling objective, where 20\% of the ICD-10 codes in each sequence are masked out and the model needs to predict the original ICD-10 code of the masked token relying only on the surrounding context of codes. We increased the mask percentage from the standard 15\% to 20\% based on \citep{Wettig:2022aa} and evaluated the model using perplexity values during training. See Table \ref{tab:pretraining} for hyperparameter choices for pre-training all models.

\subsection{Modeling Medical Text}
% Compare/contrast to ClinicalBERT etc here?
Since medical notes commonly contain more than 512 tokens, it was essential to develop models for medical text that can support much longer sequences (Figure \ref{fig:method}B). We trained different models for medical text which support document lengths of up to 8192 tokens. We used the BioLM \citep{Lewis:2020aa} \textit{RoBERTa-base-PM-M3-Voc-distill-align} as the starting model checkpoint. These models were pre-trained on text in PubMed, PMC, and MIMIC-III with a byte pair encoding vocabulary learned from PubMed. We converted them to a BigBird model to handle long sequences using the method presented in \citep{Beltagy:2020aa}. In short, we repeatedly copied over 512-position embeddings and pre-trained the model on longer text using the MIMIC-III dataset before applying additional training objectives. We refer to this model trained solely on the MIMIC-III MLM task as \textit{NoteLM}. 

\begin{table*}[h!]
 % The first argument is the label.
 % The caption goes in the second argument, and the table contents
 % go in the third argument.
\floatconts
  {tab:mimic-50}%
  {\caption{Performance on MIMIC-III-50 dataset containing common ICD-9 codes. \textit{NoteLM} and \textit{NoteContrast} were run 5 times with different seeds to report mean and standard deviation. Performance of other methods is based on results collected from papers. %For \textit{NoteLM} and \textit{NoteContrast} models we report the average of 5 fine-tuning repeats. Note that \textit{TreeMAN – all EHR} uses additional information like lab values and medications.
  }}%
  {
  \small
  \begin{tabular}{@{}lllllc@{}}

%\begin{tabular}{@{}lllllc@{}}
\toprule
\textbf{Model} &
  \multicolumn{2}{c}{\textbf{AUC}} &
  \multicolumn{2}{c}{\textbf{F1}} &
  \multicolumn{1}{c}{\textbf{Precision}} \\
\textbf{} &
  \multicolumn{1}{c}{\textbf{Macro}} &
  \multicolumn{1}{c}{\textbf{Micro}} &
  \multicolumn{1}{c}{\textbf{Macro}} &
  \multicolumn{1}{c}{\textbf{Micro}} &
  \multicolumn{1}{c}{\textbf{P@5}}  \\ \toprule  % \cmidrule(l){1-6}
% ICD-BigBird \citep{Michalopoulos:2022aa} & % (Michal. et al., 2022)
%   90.00 &
%   92.90 &
%   63.10 &
%   69.60 &
%   65.40 \\
JointLAAT \citep{Vu:2020aa} & % (Vu et al., 2020)
  92.5 &
  94.6 &
  66.1 &
  71.6 &
  67.1 \\
MSMN \citep{Yuan:2022aa} & % (Yuan et al., 2022)
  92.8 &
  94.7 &
  68.3 &
  72.5 &
  68.0 \\
KEPT \citep{Yang:2022aa}  & % (Li et al., 2022)
  92.6 &
  94.8 &
  68.9 &
  72.9 &
  67.3 \\
% ISD \citep{Zhou:2021aa} & % (Zhou et al., 2021)
%   93.5 &
%   94.9 &
%   67.9 &
%   71.7 &
%   68.2 \\
% TreeMAN - all EHR \citep{Liu:2022aa}  & % (Liu et al., 2022)
%   93.7 &
%   95.3 &
%   69.0 &
%   72.9 &
%   68.2 \\
ISD \citep{Zhou:2021aa} & % (Zhou et al., 2021)
  93.5 \(\pm\)0.4 &
  94.9 \(\pm\)0.1 &
  67.9 \(\pm\)0.9 &
  71.7 \(\pm\)0.3 &
  68.2 \(\pm\)0.5 \\
TreeMAN - all EHR \citep{Liu:2022aa}  & % (Liu et al., 2022)
  93.7 \(\pm\)0.2 &
  95.3 \(\pm\)0.0 &
  69.0 \(\pm\)0.2 &
  72.9 \(\pm\)0.2 &
  68.2 \(\pm\)0.1 \\
TreeMAN - Text \citep{Liu:2022aa}  & % (Liu et al., 2022)
  92.6 &
  94.5 &
  67.4 &
  71.4 &
  66.6 \\  \cmidrule{1-6}
NoteLM 4k &
  92.3 \(\pm\)0.10 &
  94.3 \(\pm\)0.07 &
  65.2 \(\pm\)0.35 &
  71.1 \(\pm\)0.22 &
  66.7 \(\pm\)0.11 \\
NoteContrast 4k &
  93.1 \(\pm\)0.09 &
  94.9 \(\pm\)0.06 &
  67.4 \(\pm\)0.28 &
  72.6 \(\pm\)0.24 &
  67.6 \(\pm\)0.17 \\
NoteContrast 8k &
  93.5 \(\pm\)0.14 &
  95.3 \(\pm\)0.06 &
  68.7 \(\pm\)0.46 &
  73.4 \(\pm\)0.18 &
  68.1 \(\pm\)0.21 \\
NoteContrast 8k ICD &
  \textbf{93.8} \(\pm\)0.04 &
  \textbf{95.4} \(\pm\)0.03 &
  \textbf{69.2} \(\pm\)0.21 &
  \textbf{73.6} \(\pm\)0.17 &
  \textbf{68.6} \(\pm\)0.18 \\
  \cmidrule{1-6}
%\end{tabular}

  \end{tabular}
  }
\end{table*}

\begin{table*}[h!]
 % The first argument is the label.
 % The caption goes in the second argument, and the table contents
 % go in the third argument.
\floatconts
  {tab:mimic-50-rare}%
  {\caption{Performance on MIMIC-III-50-rare dataset containing uncommon ICD-9 codes. All methods were run 5 times with different seeds to report mean and standard deviation. 
  }}%
  {\small
  \begin{tabular}{@{}lllllc@{}}
%\begin{tabular}{@{}lllllc@{}}
\toprule
\multicolumn{1}{c}{\multirow{2}{*}{\textbf{Model}}} &
  \multicolumn{2}{c}{\textbf{AUC}} &
  \multicolumn{2}{c}{\textbf{F1}} &
  \multirow{2}{*}{\textbf{Initialization}} \\ %\cmidrule(lr){2-5}
\multicolumn{1}{c}{} &
  \multicolumn{1}{c}{\textbf{Macro}} &
  \multicolumn{1}{c}{\textbf{Micro}} &
  \multicolumn{1}{c}{\textbf{Macro}} &
  \multicolumn{1}{c}{\textbf{Micro}} &
   \\ \toprule 
MSMN \citep{Yuan:2022aa}         & 75.35 \(\pm\) 1.32 & 77.41 \(\pm\)0.66 & 15.3 \(\pm\)2.77 & 16.65 \(\pm\)1.48 & \multirow{6}{*}{Pre-trained} \\
KEPT \citep{Yang:2022aa} & 79.39 \(\pm\)1.47 & 80.66 \(\pm\)1.41 & 24.61 \(\pm\)2.84 & 23.32 \(\pm\)2.15 &                                \\
NoteLM 4k                        & 82.04 \(\pm\)2.01 & 81.93 \(\pm\)1.51 & 24.89 \(\pm\)7.31 & 26.38 \(\pm\)6.11 &                                \\
NoteContrast 4k                  & \textbf{86.86} \(\pm\)1.02 & 86.45 \(\pm\)0.89 & 36.76 \(\pm\)2.56 & 36.25 \(\pm\)5.77 &                         \\
NoteContrast 8k                  & 85.65 \(\pm\)1.16        & \textbf{87.13} \(\pm\)1.23 & 38.15 \(\pm\)3.55 & 39.88 \(\pm\)2.44 &                \\
NoteContrast 8k ICD              & 85.70 \(\pm\)0.49 & 86.72 \(\pm\)1.12 & \textbf{39.08} \(\pm\)2.15 & \textbf{41.84} \(\pm\)1.56 &                 \\ \cmidrule{1-6}
MSMN \citep{Yuan:2022aa}         & 58.95 \(\pm\)4.16 & 58.9 \(\pm\)4.6 & 3.54 \(\pm\)2.18  & 5.48 \(\pm\)1.21 & \multirow{4}{*}{MIMIC-III-50} \\
KEPT \citep{Yang:2022aa} & 82.30 \(\pm\)1.73 & 83.66 \(\pm\)1.51 & 28.94 \(\pm\)1.04 & 31.43 \(\pm\)1.3 &                                  \\
NoteContrast 8k                  & 88.40 \(\pm\)0.33 & \textbf{90.01} \(\pm\)0.6 & 39.75 \(\pm\)1.48 & \textbf{43.3} \(\pm\)1.5 &                             \\
NoteContrast 8k ICD              & \textbf{88.92} \(\pm\)1.23 & 89.9 \(\pm\)0.65 & \textbf{40.26} \(\pm\)2.96 & 42.64 \(\pm\)1.97  &       \\ \cmidrule{1-6}
%\end{tabular}
  \end{tabular}
  }
\end{table*}

\subsection{Contrastive Training}
% TODO contrast our approach with KEPT (Yang) and Mullenbach here?
For the contrastive training, we used pairs of medical notes and their corresponding ICD-10 codes as positive pairs, and all other pairs as negatives. During training, we sampled batches of pairs and trained the model to predict which of the (text, ICD-10 code) pairs match across the batch (Figure \ref{fig:method}C). This was done by jointly training the text and diagnostic encoders to maximize the cosine similarity between the text and code embeddings of the positive pairs while minimizing that of the incorrect pairs in the batch. We used the pre-trained transformer models described above to encode the text and ICD-10 diagnostic code sequence and used the hidden state of the CLS token to embed the ICD-10 sequence and the medical note. Compared to the diagnostic model pre-training, which used sequences spanning multiple medical encounters, we used only a single note and diagnostic codes from the same clinical encounter, not any past or future diagnostic codes, for the contrastive training step. 
The model representations are then projected into a multi-modal embedding space ($T_N$ and $D_N$), and the InfoNCE loss \citep{Oord:2018aa} was computed as contrastive loss among positive and negative text-ICD and ICD-text pairs. In addition to the contrastive loss, we included the masked language modeling objective for the text model during training. This helped maintain existing textual properties while contrastively learning diagnostic properties. The contrastive and masked language losses were combined and weighted using uncertainty weighting \citep{Kendall:2018aa}.
We trained three model versions to be able to compare performance on downstream tasks. We first trained the \textit{NoteContrast 4k} model, which supports documents of length 4096. We started the training of this model using the NoteLM weights. We trained the model to minimize the weighted contrastive-ICD and MLM loss for 10,000 steps with a batch size of 64 and 16 gradient accumulation steps. Then, we converted the NoteContrast 4k model to handle sequences of length 8192 and trained the \textit{NoteContrast 8k} model using the same loss for 10,000 steps with a batch size of 32 and 32 gradient accumulation steps. The \textit{NoteContrast 8k ICD} model was based on the NoteContrast 8k model with additional fine-tuning using the contrastive loss objective. Here, the textual description for each ICD-10 diagnosis was treated as a very short medical note, and the associated ICD-10 code was considered a "sequence" of length 1 (Figure \ref{fig:method}D). This step brought relevant codes/text closer and separated dissimilar codes/text, improving downstream task performance.

\subsection{Prompt-based fine-tuning}

We followed the prompt-based fine-tuning approach from \cite{Yang:2022aa} for the ICD-9 coding task. Prompt-based fine-tuning is an alternative approach to multi-label classification where the multi-label classification task is reformulated as a cloze task. Each label is assigned a prompt template as shown in Figure \ref{fig:method}E and the model fills the MASK token to indicate the presence or absence of the label in the multi-label space.

\subsection{Pre-training Data}

\begin{table*}[h!]
 % The first argument is the label.
 % The caption goes in the second argument, and the table contents
 % go in the third argument.
\floatconts
  {tab:mimic-50-full}%
  {\caption{Performance on MIMIC-III-full dataset, when using \textit{NoteContrast} as a re-ranker of the top 300 \textit{MSMN} predictions.  Our model was run 5 times to report mean and standard deviation.
  }}%
  {\small
  \begin{tabular}{lllll}    
    \toprule
    \multicolumn{1}{c}{\multirow{2}{*}{\textbf{Model}}} &
      \multicolumn{2}{c}{\textbf{F1}} &
      \multicolumn{2}{c}{\textbf{Precision}} \\
    \multicolumn{1}{c}{} &
      \multicolumn{1}{c}{\textbf{Macro}} &
      \multicolumn{1}{c}{\textbf{Micro}} &
      \multicolumn{1}{c}{\textbf{@8}} &
      \multicolumn{1}{c}{\textbf{@15}}
       \\ \toprule 
    JointLAAT \citep{Vu:2020aa}       & 10.7        & 57.5        & 73.5        & 59          \\
    ISD \citep{Zhou:2021aa}           & 11.90 \(\pm\)0.2 & 55.90 \(\pm\)0.2 & 74.50 \(\pm\)0.1 & -          \\
    MSMN \citep{Yuan:2022aa}          & 10.3 \(\pm\)0.3  & 58.2 \(\pm\)0.4  & 74.9 \(\pm\)0.3  & 59.5 \(\pm\)0.1  \\
    PLM-ICD \citep{Huang:2022aa}      & 10.40 \(\pm\)0.1 & 59.80 \(\pm\)0.3 & 77.10 \(\pm\)0.2 & 61.30 \(\pm\)0.1 \\
    KEPT \citep{Yang:2022aa}  & 11.8 \(\pm\)0.4  & 59.9 \(\pm\)0.5  & 77.1 \(\pm\)0.3  & 61.5 \(\pm\)0.2  \\
    DiscNet+RE \citep{Zhang:2022aa}   & \textbf{14}          & 58.8        & 76.5        & 61.4        \\
    NoteContrast 8k ICD & 11.9 \(\pm\)0.3  & \textbf{60.7} \(\pm\)0.03 & \textbf{77.8} \(\pm\)0.1  & \textbf{62.2} \(\pm\)0.1 \\
    \cmidrule{1-5}
    %\end{tabular}
\end{tabular}
}
\end{table*}

\paragraph{ICD-10 diagnostic codes.}
% TODO removed identifier MassGeneralBrigham
A RoBERTa-style model for ICD-10 sequences was trained on hospital encounters with ICD-10 codes, consisting of nearly 60 million real-world hospital encounters from 1.5 million patients. We first prepared a sequence for each patient containing all available ICD-10 diagnostic codes, leading to 1.5 million sequences of varying lengths. We randomly selected an encounter to be the “current encounter” and by changing the current encounter, generated 5 sequences per patient that contained the same sequence of codes, but different relative position and token type values, resulting in a final dataset of 7.5 million sequences. 

\paragraph{Medical notes and contrastive training.}

The NoteLM and NoteContrast models were pre-trained on medical notes from the MIMIC-III dataset \citep{Johnson:2016aa}, a collection of medical notes from over 40,000 patients. We used nearly 2 million notes for the MLM pre-training and around 50,000 notes for the contrastive language-diagnostic pre-training. Refer to Appendix \ref{appendix:mimic-train} for more details. For better clinical utility, we trained our diagnostic code model with a vocabulary of ICD-10 codes and translated ICD9 codes to ICD-10 codes\footnote{ICD-9 to ICD-10 mapping: \url{https://github.com/AtlasCUMC/ICD-10-ICD9-codes-conversion/blob/master/ICD_9_10_d_v1.1.csv}}. To resolve ambiguous mappings, we selected an ICD10 code at random. We applied two pre-processing steps where we removed all de-identification placeholders and stripped extra white space. The NoteContrast 8k ICD model was also fine-tuned on textual descriptions of ICD-10 codes from the python package \textit{icd10-cm v0.0.5}.

\section{Experiments}

We conducted experiments with a series of NoteContrast models capable of handling document lengths of up to 4096 and 8192 tokens. Experimental results on the MIMIC-50, MIMIC-rare50, and MIMIC-III-full evaluations are shown in Tables \ref{tab:mimic-50}, \ref{tab:mimic-50-rare}, \ref{tab:mimic-50-full}. Our contrastive pre-training method demonstrated improved performance on most metrics in downstream ICD classification compared to the standard masked language modeling objective and existing state-of-the-art models for these tasks, including ISD, TreeMAN and KEPTLongformer \citep{Zhou:2021aa, Liu:2022aa, Yang:2022aa}. It is important to note that our models were pre-trained on ICD-10 codes (i.e., have not seen ICD-9 codes during pre-training), and later fine-tuned for MIMIC-50, MIMIC-rare50 and MIMIC-III-full tasks with textual ICD-9 descriptions in the prompt based fine-tuning paradigm. Despite the ICD system change from pre-training to fine tuning, our model outperforms prior approaches developed specifically for ICD-9 coding. Implementation details are discussed in Appendix \ref{appendix:implement} and Tables \ref{tab:finetune_50}, \ref{tab:finetune_rare50} and \ref{tab:finetune_full} list hyperparameter choices selected based on dev set performance for the MIMIC-III tasks.

\subsection{Dataset}

We trained and evaluated our models using de-identified discharge summaries from the MIMIC-III dataset \citep{Johnson:2016aa}, which has been widely adopted for benchmarking ICD-9 coding tasks. To allow comparison with other approaches, we adopted multiple tasks based on the MIMIC-III dataset: MIMIC-III-50, MIMIC-III-rare50, and MIMIC-III-full as previously described \citep{Yang:2022aa}. The creation, size and preprocessing details of the datasets can be found in Appendix \ref{appendix:finetune}

\subsection{Metrics}

We report micro and macro averaged F1 scores, micro and macro averaged AUC scores, precision at K (K = \{5, 8, 15\}), and recall at K (K = \{8, 15\}). All experiments were repeated 5 times with different random seeds (including model fine-tuning), and we present mean test results and standard deviation unless otherwise specified. The best thresholds for classification and computing precision, recall, and F1 were selected using the dev set for each task.

\subsection{Results}

% \begin{figure}[htbp]
%  % Caption and label go in the first argument and the figure contents
%  % go in the second argument
% \floatconts
%   {fig:nodes}
%   {\caption{Example Image}}
%   {\includegraphics[width=0.5\linewidth]{images/nodes}}
% \end{figure}

\begin{figure*}[h!]
\floatconts
  {figure:embedding}
  {\caption{Comparing outputs of the ICD-10 diagnosis model and text model before and after contrastive pre-training (left and right panels, respectively). We used a sample of 5000 ICD-10 codes as input to the ICD-10 encoder model, and textual descriptions of each code as input to the text model. For visualization, 768-dimensional outputs were projected using UMAP into two dimensions, scaled, and rotated using the Procrustes transformation. The contrastive pre-training step aligns corresponding output vectors in both models, leading to a more finely resolved embedding of all diagnoses.}}
  {\includegraphics[width=0.8\textwidth]{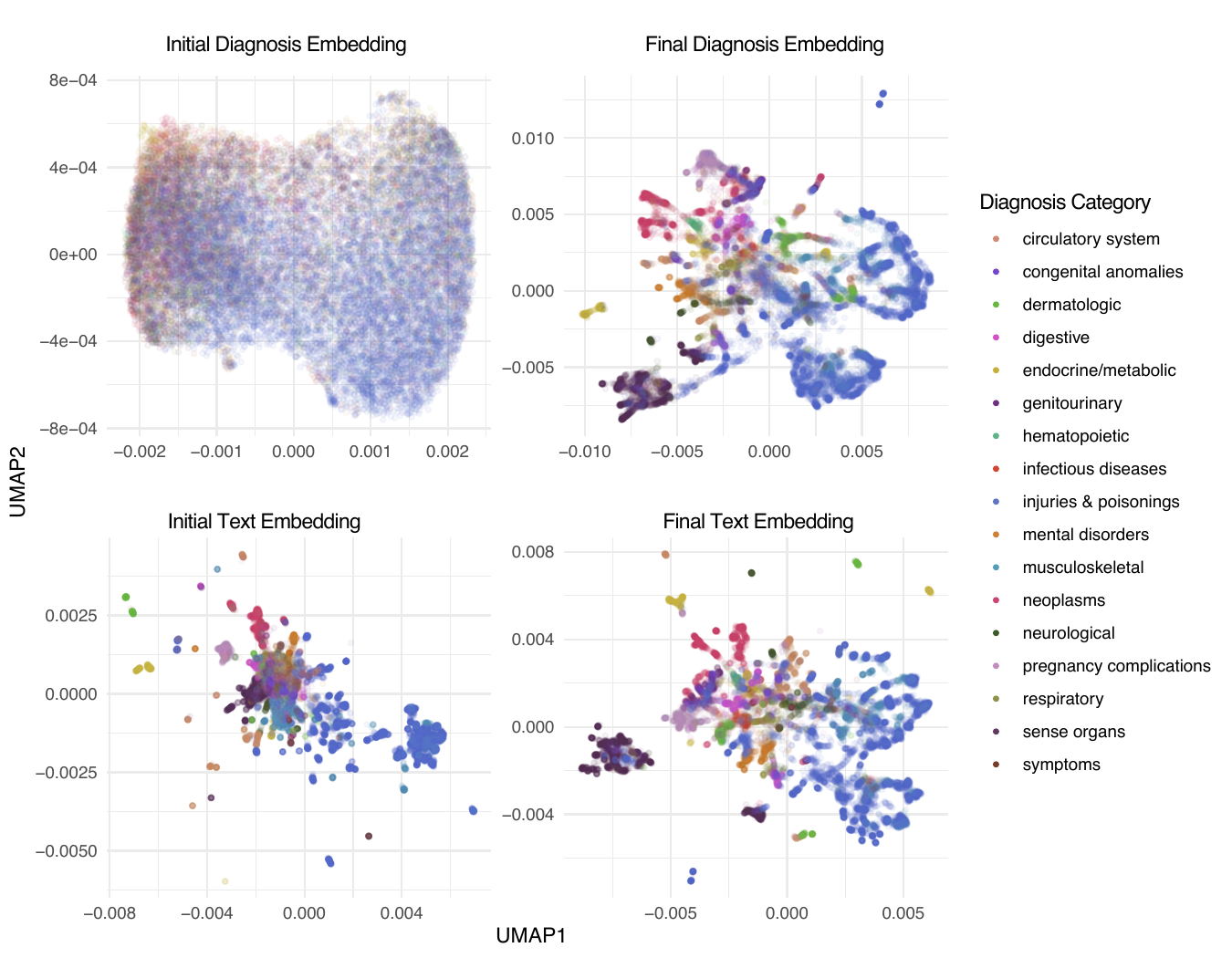}}
\end{figure*}

In the \textbf{MIMIC-III-50 task}, shown in Table \ref{tab:mimic-50}, the \textit{NoteContrast 8k ICD} model achieved a macro-AUC of 93.8 (+0.3), micro-AUC of 95.4 (+0.5), macro-F1 of 69.2 (+0.3), micro-F1 of 73.6 (+0.7), and precision@5 of 68.6 (+0.4). Numbers in parentheses show differences to prior best results. We excluded \textit{TreeMAN - all EHR} from direct comparisons, since it uses text and structured data (e.g., lab results and medications), giving it additional information compared to the other methods. Under the \textbf{MIMIC-III-rare50} setting, shown in Table \ref{tab:mimic-50-rare}, the best performance was achieved by a \textit{NoteContrast 8k ICD} model which had previously been fine-tuned on the MIMIC-III-50 task: macro-AUC of 88.92 (+6.62), micro-AUC of 89.90 (+6.24), macro-F1 of 40.26 (+11.32), micro-F1 of 42.64 (11.21). This approach performed better than fine-tuning the NoteContrast 8k ICD model on the MIMIC-III-rare50 dataset alone, which yielded macro-AUC of 85.70 (+6.31), micro-AUC of 86.72 (+6.06), macro-F1 of 39.08 (+14.47), micro-F1 of 41.84 (+18.52). For the \textbf{MIMIC-III-full} task, shown in Table \ref{tab:mimic-50-full}, using the NoteContrast 8k ICD model to re-rank the top 300 candidate codes from MSMN improved over the previous state-of-the-art method for most metrics. We achieved macro-F1 of 11.9 (-2.1), micro-F1 of 60.7 (+0.8), precision@8 of 77.8 (+0.7), precision@15 of 62.2 (+0.7), recall@8 of 41.1 (+0.4), and recall@15 of 58.3 (+0.9). 
In agreement with prior work, we observed that prompt-based fine-tuning improved performance over traditional multi-label classification \citep{Yang:2022aa}.

\section{Discussion}

Our final NoteContrast 8k ICD model combines a language model for clinical notes, contrastive training, and extends the document length to 8192 tokens. Below we compare different model versions to assess impact of each component.

\paragraph{Long-document language models offer a strong foundation for diagnostic coding of clinical notes.}
The NoteLM model, trained on 2 million notes, %, based on BioLM, is a long-document BigBird model trained on 2 million notes without diagnostic codes. This model
demonstrated comparable performance to earlier approaches such as ICD-BigBird, JointLAAT, and MSMN in the MIMIC-III-50 and MIMIC-III-rare50 tasks, without pre-training objectives related to diagnostic coding (Tables \ref{tab:mimic-50} and \ref{tab:mimic-50-rare}).

\paragraph{Diagnostic representation learning improved classification, especially for rare diagnoses. }
By incorporating a contrastive training objective between sequences of diagnostic codes and medical notes, we observed performance improvements across all evaluation tasks. Rare codes in particular (MIMIC-III-rare50, Table \ref{tab:mimic-50-rare}), had strong improvements compared to language model pre-training alone or prior work based on a hierarchical triplet-loss (e.g., KEPT). For example, contrastive training improved macro AUC by 4.82 points and macro F1 by 11.87 points for the NoteContrast 4k model compared to the NoteLM 4k model. Scaling the model from 4096 to 8192 tokens, combined with contrastive pre-training, further enhanced classification performance. The NoteContrast 8k ICD model with MIMIC-50 finetuning improved macro AUC by 6.88 and macro F1 by 15.37 points compared to the NoteLM 4k model. 

\paragraph{Contrastive training aligns outputs from diagnostic code and text models.}
 Joint training of text and diagnostic code models enabled alignment of their output vectors by maximizing the cosine similarity of positive pairs and minimizing that of incorrect pairs. Initially, the 2D UMAP projections of diagnostic code embeddings and their corresponding textual descriptions were dissimilar (Figure \ref{figure:embedding}, left panels). While the text model formed clusters corresponding to high-level diagnostic categories, the UMAP embedding primarily separated injury-related codes (light blue) from other major diagnostic categories (Figure \ref{figure:embedding}, bottom left). Conversely, the diagnostic code model assigned similar outputs to almost all non-injury codes and primarily separated different types of injuries (Figure \ref{figure:embedding}, top left). However, after contrastive pre-training, embeddings of diagnostic codes and text were similar, resulting in overall better resolution when comparing diagnoses of different major categories (Figure \ref{figure:embedding}, right panels).

\subsection{Limitations}

There are several limitations to consider in our study. Firstly, bias in the training data could affect the generalizability of our findings. The MIMIC-III dataset was sourced from a single medical institution and does not fully represent patient populations and healthcare practices found in other settings. Secondly, the long tail issue in the frequency of ICD codes poses a challenge. Our model's representations for very rare ICD codes may not be accurate, as the training data did not contain many examples for these codes. Due to vocabulary cutoff during pre-training of the ICD10 diagnosis model, certain rare codes are likely missing from our diagnostic code model, limiting the coverage of our model. Another limitation is the relatively small batch size used for contrastive learning due to computational resource limits. While our model showed promising results with a batch size of up to 64 notes and diagnostic code sequences, other contrastive models have been pre-trained on much larger batch sizes and datasets (e.g., 32k pairs of images and text in CLIP). The limited batch size might impact the stability and convergence of the training process. Lastly, there are differences in the number and type of notes used for the MLM and contrastive training steps. While we used 2 million notes of multiple types, such as progress notes, radiology reports, and discharge summaries for MLM training, we used only 50,000 discharge summaries with corresponding ICD codes for contrastive training. This could introduce biases and limitations in our model when deployed on notes other than discharge summaries.

Using real-world clinical datasets, ideally incorporating multiple institutions from different geographic regions would yield much larger collections of medical notes and diagnostic codes, improving performance for rare diagnoses and better overall generalizability.

\section{Conclusion}

The InfoNCE objective has been related to maximization of mutual information between different modalities of the same concept \citep{Oord:2018aa}, which is a powerful approach to utilize large collections of weakly labeled data. In this work, we built contextual embeddings of diagnoses based on their occurrence in a real-world data set, which has the potential to capture co-morbidities and temporally related diagnoses that occur in clinical settings. By aligning domain-specific text models with diagnostic code representations in a contrastive pre-training step, we were able to better annotate medical notes with diagnostic codes. The improvement was especially strong for infrequently occurring diagnoses, where our approach significantly improved over prior work that use fixed hierarchical sources of biomedical knowledge such as UMLS or ICD-9/ICD-10 ontologies. While this work focused on diagnostic coding of medical notes, data-driven contrastive pre-training can offer a powerful framework for other types of biomedical data with noisy labels.

\acks{

This work was supported by One Brave Idea, cofounded by the American Heart Association and Verily with significant support from AstraZeneca and pillar support from Quest Diagnostics (to C.A.M. and R.C.D.). M.H. is supported by the Drs. Tobia and Morton Mower Science Innovation Fund Fellowship.

R.C.D. was supported by grants from the National Institutes of Health and the American Heart Association (One Brave Idea, Apple Heart and Movement Study) and is a co-founder of Atman Health. C.A.M. is supported by grants from the National Institutes of Health and the American Heart Association (One Brave Idea, Apple Heart and Movement Study), is a consultant for Bayer, Biosymetrics, Clarify Health, Dewpoint Therapeutics, Dinaqor, Dr. Evidence, Foresite Labs, Insmed, Pfizer and Platform Life Sciences, and is a co-founder of Atman Health. All other authors report no competing interests.

}

\bibliography{kailas23}

\begin{thebibliography}{45}
\providecommand{\natexlab}[1]{#1}
\providecommand{\url}[1]{\texttt{#1}}
\expandafter\ifx\csname urlstyle\endcsname\relax
  \providecommand{\doi}[1]{doi: #1}\else
  \providecommand{\doi}{doi: \begingroup \urlstyle{rm}\Url}\fi

\bibitem[Alsentzer et~al.(2019)Alsentzer, Murphy, Boag, Weng, Jindi, Naumann, and McDermott]{Alsentzer:2019aa}
Emily Alsentzer, John Murphy, William Boag, Wei-Hung Weng, Di~Jindi, Tristan Naumann, and Matthew McDermott.
\newblock Publicly available clinical {BERT} embeddings.
\newblock In \emph{Proceedings of the 2nd Clinical Natural Language Processing Workshop}, pages 72--78, Minneapolis, Minnesota, USA, 2019. Association for Computational Linguistics.
\newblock \doi{10.18653/v1/W19-1909}.
\newblock URL \url{https://aclanthology.org/W19-1909}.

\bibitem[Beltagy et~al.(2019)Beltagy, Lo, and Cohan]{Beltagy:2019aa}
Iz~Beltagy, Kyle Lo, and Arman Cohan.
\newblock {S}ci{BERT}: A pretrained language model for scientific text.
\newblock In \emph{Proc. of EMNLP}, pages 3615--3620, Hong Kong, China, 2019. Association for Computational Linguistics.
\newblock \doi{10.18653/v1/D19-1371}.
\newblock URL \url{https://aclanthology.org/D19-1371}.

\bibitem[Beltagy et~al.(2020)Beltagy, Peters, and Cohan]{Beltagy:2020aa}
Iz~Beltagy, Matthew~E Peters, and Arman Cohan.
\newblock {Longformer: The Long-Document Transformer}.
\newblock \emph{arXiv}, 2020.
\newblock \doi{10.48550/arxiv.2004.05150}.

\bibitem[Cao et~al.(2020)Cao, Chen, Liu, Zhao, Liu, and Chong]{Cao:2020aa}
Pengfei Cao, Yubo Chen, Kang Liu, Jun Zhao, Shengping Liu, and Weifeng Chong.
\newblock {H}yper{C}ore: Hyperbolic and co-graph representation for automatic {ICD} coding.
\newblock In \emph{Proc. of ACL}, pages 3105--3114, Online, 2020. Association for Computational Linguistics.
\newblock \doi{10.18653/v1/2020.acl-main.282}.
\newblock URL \url{https://aclanthology.org/2020.acl-main.282}.

\bibitem[Chen et~al.(2020)Chen, Kornblith, Norouzi, and Hinton]{Chen:2020aa}
Ting Chen, Simon Kornblith, Mohammad Norouzi, and Geoffrey~E. Hinton.
\newblock A simple framework for contrastive learning of visual representations.
\newblock In \emph{Proc. of ICML}, volume 119 of \emph{Proceedings of Machine Learning Research}, pages 1597--1607. {PMLR}, 2020.
\newblock URL \url{http://proceedings.mlr.press/v119/chen20j.html}.

\bibitem[Devlin et~al.(2019)Devlin, Chang, Lee, and Toutanova]{Devlin:2019aa}
Jacob Devlin, Ming-Wei Chang, Kenton Lee, and Kristina Toutanova.
\newblock {BERT}: Pre-training of deep bidirectional transformers for language understanding.
\newblock In \emph{Proc. of NAACL-HLT}, pages 4171--4186, Minneapolis, Minnesota, 2019. Association for Computational Linguistics.
\newblock \doi{10.18653/v1/N19-1423}.
\newblock URL \url{https://aclanthology.org/N19-1423}.

\bibitem[Huang et~al.(2022)Huang, Tsai, and Chen]{Huang:2022aa}
Chao-Wei Huang, Shang-Chi Tsai, and Yun-Nung Chen.
\newblock {PLM}-{ICD}: Automatic {ICD} coding with pretrained language models.
\newblock In \emph{Proceedings of the 4th Clinical Natural Language Processing Workshop}, pages 10--20, Seattle, WA, 2022. Association for Computational Linguistics.
\newblock \doi{10.18653/v1/2022.clinicalnlp-1.2}.
\newblock URL \url{https://aclanthology.org/2022.clinicalnlp-1.2}.

\bibitem[Huang et~al.(2021)Huang, Shen, Lungren, and Yeung]{Huang:2021aa}
Shih{-}Cheng Huang, Liyue Shen, Matthew~P. Lungren, and Serena Yeung.
\newblock Gloria: {A} multimodal global-local representation learning framework for label-efficient medical image recognition.
\newblock In \emph{2021 {IEEE/CVF} International Conference on Computer Vision, {ICCV} 2021, Montreal, QC, Canada, October 10-17, 2021}, pages 3922--3931. {IEEE}, 2021.
\newblock \doi{10.1109/ICCV48922.2021.00391}.
\newblock URL \url{https://doi.org/10.1109/ICCV48922.2021.00391}.

\bibitem[Izacard et~al.(2021)Izacard, Caron, Hosseini, Riedel, Bojanowski, Joulin, and Grave]{Izacard:2021aa}
Gautier Izacard, Mathilde Caron, Lucas Hosseini, Sebastian Riedel, Piotr Bojanowski, Armand Joulin, and Edouard Grave.
\newblock {Unsupervised Dense Information Retrieval with Contrastive Learning}.
\newblock \emph{arXiv}, 2021.
\newblock \doi{10.48550/arxiv.2112.09118}.

\bibitem[Ji et~al.(2020)Ji, Cambria, and Marttinen]{Ji:2020aa}
Shaoxiong Ji, Erik Cambria, and Pekka Marttinen.
\newblock Dilated convolutional attention network for medical code assignment from clinical text.
\newblock In \emph{Proceedings of the 3rd Clinical Natural Language Processing Workshop}, pages 73--78, Online, 2020. Association for Computational Linguistics.
\newblock \doi{10.18653/v1/2020.clinicalnlp-1.8}.
\newblock URL \url{https://aclanthology.org/2020.clinicalnlp-1.8}.

\bibitem[Jia et~al.(2021)Jia, Yang, Xia, Chen, Parekh, Pham, Le, Sung, Li, and Duerig]{Jia:2021aa}
Chao Jia, Yinfei Yang, Ye~Xia, Yi{-}Ting Chen, Zarana Parekh, Hieu Pham, Quoc~V. Le, Yun{-}Hsuan Sung, Zhen Li, and Tom Duerig.
\newblock Scaling up visual and vision-language representation learning with noisy text supervision.
\newblock In Marina Meila and Tong Zhang, editors, \emph{Proc. of ICML}, volume 139 of \emph{Proceedings of Machine Learning Research}, pages 4904--4916. {PMLR}, 2021.
\newblock URL \url{http://proceedings.mlr.press/v139/jia21b.html}.

\bibitem[Johnson et~al.(2016)Johnson, Pollard, Shen, Lehman, Feng, Ghassemi, Moody, Szolovits, Celi, and Mark]{Johnson:2016aa}
Alistair~E.W. Johnson, Tom~J. Pollard, Lu~Shen, Li-wei~H. Lehman, Mengling Feng, Mohammad Ghassemi, Benjamin Moody, Peter Szolovits, Leo~Anthony Celi, and Roger~G. Mark.
\newblock {MIMIC-III, a freely accessible critical care database}.
\newblock \emph{Scientific Data}, 3:\penalty0 sdata201635, 2016.
\newblock ISSN 2052-4463.
\newblock \doi{10.1038/sdata.2016.35}.

\bibitem[Kendall et~al.(2018)Kendall, Gal, and Cipolla]{Kendall:2018aa}
Alex Kendall, Yarin Gal, and Roberto Cipolla.
\newblock Multi-task learning using uncertainty to weigh losses for scene geometry and semantics.
\newblock In \emph{2018 {IEEE} Conference on Computer Vision and Pattern Recognition, {CVPR} 2018, Salt Lake City, UT, USA, June 18-22, 2018}, pages 7482--7491. {IEEE} Computer Society, 2018.
\newblock \doi{10.1109/CVPR.2018.00781}.
\newblock URL \url{http://openaccess.thecvf.com/content\_cvpr\_2018/html/Kendall\_Multi-Task\_Learning\_Using\_CVPR\_2018\_paper.html}.

\bibitem[Kingma and Ba(2015)]{Kingma:2014}
Diederik~P. Kingma and Jimmy Ba.
\newblock Adam: {A} method for stochastic optimization.
\newblock In Yoshua Bengio and Yann LeCun, editors, \emph{Proc. of ICLR}, 2015.
\newblock URL \url{http://arxiv.org/abs/1412.6980}.

\bibitem[Lee et~al.(2019)Lee, Yoon, Kim, Kim, Kim, So, and Kang]{Lee:2019aa}
Jinhyuk Lee, Wonjin Yoon, Sungdong Kim, Donghyeon Kim, Sunkyu Kim, Chan~Ho So, and Jaewoo Kang.
\newblock {BioBERT: a pre-trained biomedical language representation model for biomedical text mining}.
\newblock \emph{Bioinformatics}, 36\penalty0 (4):\penalty0 1234--1240, 2019.
\newblock ISSN 1367-4803.
\newblock \doi{10.1093/bioinformatics/btz682}.

\bibitem[Lewis et~al.(2020)Lewis, Ott, Du, and Stoyanov]{Lewis:2020aa}
Patrick Lewis, Myle Ott, Jingfei Du, and Veselin Stoyanov.
\newblock Pretrained language models for biomedical and clinical tasks: Understanding and extending the state-of-the-art.
\newblock In \emph{Proceedings of the 3rd Clinical Natural Language Processing Workshop}, pages 146--157, Online, 2020. Association for Computational Linguistics.
\newblock \doi{10.18653/v1/2020.clinicalnlp-1.17}.
\newblock URL \url{https://aclanthology.org/2020.clinicalnlp-1.17}.

\bibitem[Li and Yu(2020)]{Li:2020ab}
Fei Li and Hong Yu.
\newblock {ICD} coding from clinical text using multi-filter residual convolutional neural network.
\newblock In \emph{The Thirty-Fourth {AAAI} Conference on Artificial Intelligence, {AAAI} 2020, The Thirty-Second Innovative Applications of Artificial Intelligence Conference, {IAAI} 2020, The Tenth {AAAI} Symposium on Educational Advances in Artificial Intelligence, {EAAI} 2020, New York, NY, USA, February 7-12, 2020}, pages 8180--8187. {AAAI} Press, 2020.
\newblock URL \url{https://aaai.org/ojs/index.php/AAAI/article/view/6331}.

\bibitem[Li et~al.(2020)Li, Rao, Solares, Hassaine, Ramakrishnan, Canoy, Zhu, Rahimi, and Salimi-Khorshidi]{Li:2020aa}
Yikuan Li, Shishir Rao, Jos{\'e} Roberto~Ayala Solares, Abdelaali Hassaine, Rema Ramakrishnan, Dexter Canoy, Yajie Zhu, Kazem Rahimi, and Gholamreza Salimi-Khorshidi.
\newblock {BEHRT: Transformer for Electronic Health Records}.
\newblock \emph{Scientific Reports}, 10\penalty0 (1):\penalty0 7155, 2020.
\newblock \doi{10.1038/s41598-020-62922-y}.

\bibitem[Li et~al.(2022)Li, Wehbe, Ahmad, Wang, and Luo]{Li:2022aa}
Yikuan Li, Ramsey~M Wehbe, Faraz~S Ahmad, Hanyin Wang, and Yuan Luo.
\newblock {Clinical-Longformer and Clinical-BigBird: Transformers for long clinical sequences}.
\newblock \emph{arXiv}, 2022.
\newblock \doi{10.48550/arxiv.2201.11838}.

\bibitem[Liu et~al.(2019)Liu, Ott, Goyal, Du, Joshi, Chen, Levy, Lewis, Zettlemoyer, and Stoyanov]{Liu:2019aa}
Yinhan Liu, Myle Ott, Naman Goyal, Jingfei Du, Mandar Joshi, Danqi Chen, Omer Levy, Mike Lewis, Luke Zettlemoyer, and Veselin Stoyanov.
\newblock {RoBERTa: A Robustly Optimized BERT Pretraining Approach}.
\newblock \emph{arXiv}, 2019.
\newblock \doi{10.48550/arxiv.1907.11692}.

\bibitem[Liu et~al.(2022)Liu, Liu, Wen, Zhao, Xia, and Yuan]{Liu:2022aa}
Zichen Liu, Xuyuan Liu, Yanlong Wen, Guoqing Zhao, Fen Xia, and Xiaojie Yuan.
\newblock {T}ree{MAN}: Tree-enhanced multimodal attention network for {ICD} coding.
\newblock In \emph{Proceedings of the 29th International Conference on Computational Linguistics}, pages 3054--3063, Gyeongju, Republic of Korea, 2022. International Committee on Computational Linguistics.
\newblock URL \url{https://aclanthology.org/2022.coling-1.270}.

\bibitem[Loshchilov and Hutter(2019)]{Loshchilov:2019}
Ilya Loshchilov and Frank Hutter.
\newblock Decoupled weight decay regularization.
\newblock In \emph{Proc. of ICLR}. OpenReview.net, 2019.
\newblock URL \url{https://openreview.net/forum?id=Bkg6RiCqY7}.

\bibitem[Michalopoulos et~al.(2022)Michalopoulos, Malyska, Sahar, Wong, and Chen]{Michalopoulos:2022aa}
George Michalopoulos, Michal Malyska, Nicola Sahar, Alexander Wong, and Helen Chen.
\newblock {ICDB}ig{B}ird: A contextual embedding model for {ICD} code classification.
\newblock In \emph{Proceedings of the 21st Workshop on Biomedical Language Processing}, pages 330--336, Dublin, Ireland, 2022. Association for Computational Linguistics.
\newblock \doi{10.18653/v1/2022.bionlp-1.32}.
\newblock URL \url{https://aclanthology.org/2022.bionlp-1.32}.

\bibitem[Mullenbach et~al.(2018)Mullenbach, Wiegreffe, Duke, Sun, and Eisenstein]{Mullenbach:2018aa}
James Mullenbach, Sarah Wiegreffe, Jon Duke, Jimeng Sun, and Jacob Eisenstein.
\newblock Explainable prediction of medical codes from clinical text.
\newblock In \emph{Proc. of NAACL-HLT}, pages 1101--1111, New Orleans, Louisiana, 2018. Association for Computational Linguistics.
\newblock \doi{10.18653/v1/N18-1100}.
\newblock URL \url{https://aclanthology.org/N18-1100}.

\bibitem[M{\"u}ller et~al.(2021)M{\"u}ller, Kaissis, Zou, and Rueckert]{Muller:2021aa}
Philip M{\"u}ller, Georgios Kaissis, Congyu Zou, and Daniel Rueckert.
\newblock {Joint Learning of Localized Representations from Medical Images and Reports}.
\newblock \emph{arXiv}, 2021.
\newblock \doi{10.48550/arxiv.2112.02889}.

\bibitem[Neelakantan et~al.(2022)Neelakantan, Xu, Puri, Radford, Han, Tworek, Yuan, Tezak, Kim, Hallacy, Heidecke, Shyam, Power, Nekoul, Sastry, Krueger, Schnurr, Such, Hsu, Thompson, Khan, Sherbakov, Jang, Welinder, and Weng]{Neelakantan:2022aa}
Arvind Neelakantan, Tao Xu, Raul Puri, Alec Radford, Jesse~Michael Han, Jerry Tworek, Qiming Yuan, Nikolas Tezak, Jong~Wook Kim, Chris Hallacy, Johannes Heidecke, Pranav Shyam, Boris Power, Tyna~Eloundou Nekoul, Girish Sastry, Gretchen Krueger, David Schnurr, Felipe~Petroski Such, Kenny Hsu, Madeleine Thompson, Tabarak Khan, Toki Sherbakov, Joanne Jang, Peter Welinder, and Lilian Weng.
\newblock {Text and Code Embeddings by Contrastive Pre-Training}.
\newblock \emph{arXiv}, 2022.
\newblock \doi{10.48550/arxiv.2201.10005}.

\bibitem[Oord et~al.(2018)Oord, Li, and Vinyals]{Oord:2018aa}
Aaron van~den Oord, Yazhe Li, and Oriol Vinyals.
\newblock {Representation Learning with Contrastive Predictive Coding}.
\newblock \emph{arXiv}, 2018.
\newblock \doi{10.48550/arxiv.1807.03748}.

\bibitem[Paszke et~al.(2019)Paszke, Gross, Massa, Lerer, Bradbury, Chanan, Killeen, Lin, Gimelshein, Antiga, Desmaison, K{\"{o}}pf, Yang, DeVito, Raison, Tejani, Chilamkurthy, Steiner, Fang, Bai, and Chintala]{Paszke:2019}
Adam Paszke, Sam Gross, Francisco Massa, Adam Lerer, James Bradbury, Gregory Chanan, Trevor Killeen, Zeming Lin, Natalia Gimelshein, Luca Antiga, Alban Desmaison, Andreas K{\"{o}}pf, Edward Yang, Zachary DeVito, Martin Raison, Alykhan Tejani, Sasank Chilamkurthy, Benoit Steiner, Lu~Fang, Junjie Bai, and Soumith Chintala.
\newblock Pytorch: An imperative style, high-performance deep learning library.
\newblock In Hanna~M. Wallach, Hugo Larochelle, Alina Beygelzimer, Florence d'Alch{\'{e}}{-}Buc, Emily~B. Fox, and Roman Garnett, editors, \emph{Advances in Neural Information Processing Systems 32: Annual Conference on Neural Information Processing Systems 2019, NeurIPS 2019, December 8-14, 2019, Vancouver, BC, Canada}, pages 8024--8035, 2019.
\newblock URL \url{https://proceedings.neurips.cc/paper/2019/hash/bdbca288fee7f92f2bfa9f7012727740-Abstract.html}.

\bibitem[Radford et~al.(2021)Radford, Kim, Hallacy, Ramesh, Goh, Agarwal, Sastry, Askell, Mishkin, Clark, Krueger, and Sutskever]{Radford:2021aa}
Alec Radford, Jong~Wook Kim, Chris Hallacy, Aditya Ramesh, Gabriel Goh, Sandhini Agarwal, Girish Sastry, Amanda Askell, Pamela Mishkin, Jack Clark, Gretchen Krueger, and Ilya Sutskever.
\newblock Learning transferable visual models from natural language supervision.
\newblock In Marina Meila and Tong Zhang, editors, \emph{Proc. of ICML}, volume 139 of \emph{Proceedings of Machine Learning Research}, pages 8748--8763. {PMLR}, 2021.
\newblock URL \url{http://proceedings.mlr.press/v139/radford21a.html}.

\bibitem[Rasmy et~al.(2021)Rasmy, Xiang, Xie, Tao, and Zhi]{Rasmy:2021aa}
Laila Rasmy, Yang Xiang, Ziqian Xie, Cui Tao, and Degui Zhi.
\newblock {Med-BERT: pretrained contextualized embeddings on large-scale structured electronic health records for disease prediction}.
\newblock \emph{npj Digital Medicine}, 4\penalty0 (1):\penalty0 86, 2021.
\newblock \doi{10.1038/s41746-021-00455-y}.

\bibitem[Shang et~al.(2019)Shang, Ma, Xiao, and Sun]{Shang:2019aa}
Junyuan Shang, Tengfei Ma, Cao Xiao, and Jimeng Sun.
\newblock Pre-training of graph augmented transformers for medication recommendation.
\newblock In Sarit Kraus, editor, \emph{Proceedings of the Twenty-Eighth International Joint Conference on Artificial Intelligence, {IJCAI} 2019, Macao, China, August 10-16, 2019}, pages 5953--5959. ijcai.org, 2019.
\newblock \doi{10.24963/ijcai.2019/825}.
\newblock URL \url{https://doi.org/10.24963/ijcai.2019/825}.

\bibitem[Vaswani et~al.(2017)Vaswani, Shazeer, Parmar, Uszkoreit, Jones, Gomez, Kaiser, and Polosukhin]{Vaswani:2017aa}
Ashish Vaswani, Noam Shazeer, Niki Parmar, Jakob Uszkoreit, Llion Jones, Aidan~N. Gomez, Lukasz Kaiser, and Illia Polosukhin.
\newblock Attention is all you need.
\newblock In Isabelle Guyon, Ulrike von Luxburg, Samy Bengio, Hanna~M. Wallach, Rob Fergus, S.~V.~N. Vishwanathan, and Roman Garnett, editors, \emph{Advances in Neural Information Processing Systems 30: Annual Conference on Neural Information Processing Systems 2017, December 4-9, 2017, Long Beach, CA, {USA}}, pages 5998--6008, 2017.
\newblock URL \url{https://proceedings.neurips.cc/paper/2017/hash/3f5ee243547dee91fbd053c1c4a845aa-Abstract.html}.

\bibitem[Vu et~al.(2020)Vu, Nguyen, and Nguyen]{Vu:2020aa}
Thanh Vu, Dat~Quoc Nguyen, and Anthony Nguyen.
\newblock A label attention model for {ICD} coding from clinical text.
\newblock In Christian Bessiere, editor, \emph{Proceedings of the Twenty-Ninth International Joint Conference on Artificial Intelligence, {IJCAI} 2020}, pages 3335--3341. ijcai.org, 2020.
\newblock \doi{10.24963/ijcai.2020/461}.
\newblock URL \url{https://doi.org/10.24963/ijcai.2020/461}.

\bibitem[Wang et~al.(2021)Wang, Xu, Tam, Yang, and Xu]{Wang:2021aa}
Xiaosong Wang, Ziyue Xu, Leo Tam, Dong Yang, and Daguang Xu.
\newblock {Self-supervised Image-text Pre-training With Mixed Data In Chest X-rays}.
\newblock \emph{arXiv}, 2021.
\newblock \doi{10.48550/arxiv.2103.16022}.

\bibitem[Wang et~al.(2022)Wang, Wu, Agarwal, and Sun]{Wang:2022aa}
Zifeng Wang, Zhenbang Wu, Dinesh Agarwal, and Jimeng Sun.
\newblock {M}ed{CLIP}: Contrastive learning from unpaired medical images and text.
\newblock In \emph{Proc. of EMNLP}, pages 3876--3887, Abu Dhabi, United Arab Emirates, 2022. Association for Computational Linguistics.
\newblock URL \url{https://aclanthology.org/2022.emnlp-main.256}.

\bibitem[Wettig et~al.(2023)Wettig, Gao, Zhong, and Chen]{Wettig:2022aa}
Alexander Wettig, Tianyu Gao, Zexuan Zhong, and Danqi Chen.
\newblock Should you mask 15{\%} in masked language modeling?
\newblock In \emph{Proceedings of the 17th Conference of the European Chapter of the Association for Computational Linguistics}, pages 2985--3000, Dubrovnik, Croatia, 2023. Association for Computational Linguistics.
\newblock URL \url{https://aclanthology.org/2023.eacl-main.217}.

\bibitem[Wolf et~al.(2020)Wolf, Debut, Sanh, Chaumond, Delangue, Moi, Cistac, Rault, Louf, Funtowicz, Davison, Shleifer, von Platen, Ma, Jernite, Plu, Xu, Le~Scao, Gugger, Drame, Lhoest, and Rush]{Wolf:2020}
Thomas Wolf, Lysandre Debut, Victor Sanh, Julien Chaumond, Clement Delangue, Anthony Moi, Pierric Cistac, Tim Rault, Remi Louf, Morgan Funtowicz, Joe Davison, Sam Shleifer, Patrick von Platen, Clara Ma, Yacine Jernite, Julien Plu, Canwen Xu, Teven Le~Scao, Sylvain Gugger, Mariama Drame, Quentin Lhoest, and Alexander Rush.
\newblock Transformers: State-of-the-art natural language processing.
\newblock In \emph{Proc. of EMNLP}, pages 38--45, Online, 2020. Association for Computational Linguistics.
\newblock \doi{10.18653/v1/2020.emnlp-demos.6}.
\newblock URL \url{https://aclanthology.org/2020.emnlp-demos.6}.

\bibitem[Xie et~al.(2019)Xie, Xiong, Yu, and Zhu]{Xie:2019aa}
Xiancheng Xie, Yun Xiong, Philip~S. Yu, and Yangyong Zhu.
\newblock {EHR} coding with multi-scale feature attention and structured knowledge graph propagation.
\newblock In Wenwu Zhu, Dacheng Tao, Xueqi Cheng, Peng Cui, Elke~A. Rundensteiner, David Carmel, Qi~He, and Jeffrey~Xu Yu, editors, \emph{Proceedings of the 28th {ACM} International Conference on Information and Knowledge Management, {CIKM} 2019, Beijing, China, November 3-7, 2019}, pages 649--658. {ACM}, 2019.
\newblock \doi{10.1145/3357384.3357897}.
\newblock URL \url{https://doi.org/10.1145/3357384.3357897}.

\bibitem[Yang et~al.(2022)Yang, Wang, Rawat, Mitra, and Yu]{Yang:2022aa}
Zhichao Yang, Shufan Wang, Bhanu Pratap~Singh Rawat, Avijit Mitra, and Hong Yu.
\newblock Knowledge injected prompt based fine-tuning for multi-label few-shot {ICD} coding.
\newblock In \emph{Findings of the Association for Computational Linguistics: EMNLP 2022}, pages 1767--1781, Abu Dhabi, United Arab Emirates, 2022. Association for Computational Linguistics.
\newblock URL \url{https://aclanthology.org/2022.findings-emnlp.127}.

\bibitem[Yuan et~al.(2022)Yuan, Tan, and Huang]{Yuan:2022aa}
Zheng Yuan, Chuanqi Tan, and Songfang Huang.
\newblock Code synonyms do matter: Multiple synonyms matching network for automatic {ICD} coding.
\newblock In \emph{Proc. of ACL}, pages 808--814, Dublin, Ireland, 2022. Association for Computational Linguistics.
\newblock \doi{10.18653/v1/2022.acl-short.91}.
\newblock URL \url{https://aclanthology.org/2022.acl-short.91}.

\bibitem[Zaheer et~al.(2020)Zaheer, Guruganesh, Dubey, Ainslie, Alberti, Onta{\~{n}}{\'{o}}n, Pham, Ravula, Wang, Yang, and Ahmed]{Zaheer:2022aa}
Manzil Zaheer, Guru Guruganesh, Kumar~Avinava Dubey, Joshua Ainslie, Chris Alberti, Santiago Onta{\~{n}}{\'{o}}n, Philip Pham, Anirudh Ravula, Qifan Wang, Li~Yang, and Amr Ahmed.
\newblock Big bird: Transformers for longer sequences.
\newblock In Hugo Larochelle, Marc'Aurelio Ranzato, Raia Hadsell, Maria{-}Florina Balcan, and Hsuan{-}Tien Lin, editors, \emph{Advances in Neural Information Processing Systems 33: Annual Conference on Neural Information Processing Systems 2020, NeurIPS 2020, December 6-12, 2020, virtual}, 2020.
\newblock URL \url{https://proceedings.neurips.cc/paper/2020/hash/c8512d142a2d849725f31a9a7a361ab9-Abstract.html}.

\bibitem[Zang and Wang(2021)]{Zang:2021aa}
Chengxi Zang and Fei Wang.
\newblock {SCEHR: Supervised Contrastive Learning for Clinical Risk Prediction using Electronic Health Records}.
\newblock \emph{2021 IEEE International Conference on Data Mining (ICDM)}, 00:\penalty0 857--866, 2021.
\newblock ISSN 1550-4786.
\newblock \doi{10.1109/icdm51629.2021.00097}.

\bibitem[Zhang et~al.(2022)Zhang, Zhang, Zhang, Sang, and Yang]{Zhang:2022aa}
Shurui Zhang, Bozheng Zhang, Fuxin Zhang, Bo~Sang, and Wanchun Yang.
\newblock Automatic {ICD} coding exploiting discourse structure and reconciled code embeddings.
\newblock In \emph{Proceedings of the 29th International Conference on Computational Linguistics}, pages 2883--2891, Gyeongju, Republic of Korea, 2022. International Committee on Computational Linguistics.
\newblock URL \url{https://aclanthology.org/2022.coling-1.254}.

\bibitem[Zhang et~al.(2020)Zhang, Jiang, Miura, Manning, and Langlotz]{Zhang:2020aa}
Yuhao Zhang, Hang Jiang, Yasuhide Miura, Christopher~D Manning, and Curtis~P Langlotz.
\newblock {Contrastive Learning of Medical Visual Representations from Paired Images and Text}.
\newblock \emph{Proceedings of the 7th Machine Learning for Healthcare Conference}, 182:\penalty0 2--25, 2020.
\newblock URL \url{https://proceedings.mlr.press/v182/zhang22a.html}.

\bibitem[Zhou et~al.(2021)Zhou, Cao, Chen, Liu, Zhao, Niu, Chong, and Liu]{Zhou:2021aa}
Tong Zhou, Pengfei Cao, Yubo Chen, Kang Liu, Jun Zhao, Kun Niu, Weifeng Chong, and Shengping Liu.
\newblock Automatic {ICD} coding via interactive shared representation networks with self-distillation mechanism.
\newblock In \emph{Proc. of ACL}, pages 5948--5957, Online, 2021. Association for Computational Linguistics.
\newblock \doi{10.18653/v1/2021.acl-long.463}.
\newblock URL \url{https://aclanthology.org/2021.acl-long.463}.

\end{thebibliography}

\appendix

\section{Appendix}\label{sec:appendix}

\subsection{Ethics Approval}

The study protocol was approved by the Mass General Brigham Institutional Review Board.

\subsection{Dataset Processing}

\subsubsection{ICD-10 diagnostic codes.} \label{appendix:icd-pre}
ICD-10 codes from real-world hospital encounters from the MassGeneral Brigham hospital system were used to build the ICD-10 sequence dataset. We excluded any patient that had less than 5 hospital encounters. For each patient, we generated 5 sequences with the same set of codes but with different "current encounters" resulting in different relative position and token type values. The final dataset consisted of 7,502,320 sequences of hospital encounters from 1,500,464 patients of which 7,447,575 were used for training and the remaining 54,745 was used for validation. All sequences belonging to a given patient are either in the train or dev set (i.e., a patient cannot have sequences in both the train and dev set). Each ICD-10 sequence on average contained 86.64 ICD-10 codes.

\subsubsection{Medical notes and contrastive training.} \label{appendix:mimic-train}
The MIMIC-III dataset \citep{Johnson:2016aa} contains 2,083,180 million de-identified notes. We removed all patients that appear in the test dataset of any evaluation task (MIMIC-50, MIMIC-rare50, and MIMIC-III-full). For masked language model (MLM) pre-training, the training set consisted of 2,059,772 notes and the dev set contained 20,036 notes. In the contrastive language diagnostic pre-training, we utilized only those notes that have diagnostic codes available. We used a training set of 47,707 notes and a dev set of 1,631 notes. All notes that were used for contrastive pre-training were discharge summaries. The pre-processing of the data for both MLM and contrastive pre-training was minimal and only included removal of all de-identification placeholders present in the MIMIC dataset and stripping of extra white spaces.

\subsubsection{Fine-tuning and evaluation} \label{appendix:finetune}
We used the steps described by \citep{Yang:2022aa} for creating the train, dev and test datasets for the MIMIC-50, MIMIC-rare50, and MIMIC-III-full tasks\footnote{MIMIC-III preprocessing code: \url{https://github.com/whaleloops/KEPT\#download--preprocess-data}}. MIMIC-50 contains instances that had at least one of the top 50 most frequent codes. MIMIC-rare50 introduced by \citep{Yang:2022aa} is built by selecting the top 50 codes with less than 10 occurrences and contains instances that have at least one of these rare codes. MIMIC-III-full includes all discharge summaries. The processing of this dataset and the resulting train, dev and test splits have been used to benchmark multiple previous approaches \citep{Mullenbach:2018aa, Vu:2020aa, Yuan:2022aa}.

\subsection{Implementation Details} \label{appendix:implement}

The ICD sequence encoder model was trained for 200K steps with a 2K batch size and took about 4 days of training time. The NoteLM and NoteContrast models were trained with a 1k batch size with varying gradient accumulation steps. NoteLM was trained for 7K steps and took about a day to train. The NoteLM model checkpoint with the lowest perplexity on the dev set was selected for all experiments. NoteContrast 4k and NoteContrast 8k were trained for 10K steps, NoteContrast 4k took about 2.5 days to train, while NoteContrast 8k took 5 days. NoteContrast 8k ICD was trained for 250 steps and took less than 5 minutes. The NoteContrast model checkpoints with the lowest contrastive loss on the dev set were used for all downstream evaluation tasks. We list the detailed hyperparameters for the pre-trained ICD and text models in Table \ref{tab:pretraining}. For the 3 downstream tasks (MIMIC-50, MIMIC-rare50, and MIMIC-III-full) we tuned the learning rate and weight decay using the dev set. The MIMIC-III-50 task took between 1-2 hours to complete training, the MIMIC-III-rare50 took between 25-50 minutes, and the MIMIC-III-full task took about 20 hours. The fine-tuning hyperparameters are listed in Tables \ref{tab:finetune_50}, \ref{tab:finetune_rare50} and \ref{tab:finetune_full} for the MIMIC-III-50, MIMIC-III-rare50 and MIMIC-III-full tasks. The macro-AUC and micro-AUC performance on the dev set was used to select the model checkpoints for evaluation in the aforementioned tasks. All models were trained and fine-tuned on a DGX-2 node with 8 A100 GPUs. We used Adam \citep{Kingma:2014} as the optimizer with weight decay \citep{Loshchilov:2019} for pre-training and fine-tuning all models. Our code is implemented based on PyTorch \citep{Paszke:2019} and Huggingface Transformers \citep{Wolf:2020} and available at \url{https://github.com/obi-ml-public/NoteContrast}. Since real-world data were used to train ICD-10 sequence models, we are unable to share the resulting model weights.

\begin{table*}[h!]
\centering
\small
\begin{tabular}{@{}llllll@{}}
\toprule
Hyper-parameter & ICD-10 Encoder & NoteLM        & NoteContrast 4k & NoteC. 8k        & NoteC. 8k ICD \\ \midrule
Base Model      & RoBERTa        & BigBird BioLM & NoteLM          & NoteC. 4k & NoteC. 8k     \\
Dropout                 & 0.1      & 0.1      & 0.1      & 0.1      & 0.1      \\
Warmup Steps            & 4000     & 1000     & 1000     & 1000     & 50       \\
Learning Rate   & 7.00E-04       & 5.00E-04      & 1.00E-04        & 7.50E-05               & 1.00E-04            \\
Device Batch Size       & 64       & 64       & 64       & 32       & 1024     \\
Gradient Accumulation   & 4        & 16       & 16       & 32       & 1        \\
Effective Batch Size    & 2048     & 1024     & 1024     & 1024     & 1024     \\
Weight Decay            & 0.01     & 0.01     & 0.1      & 0.1      & 0.1      \\
Max Steps               & 200000   & 7000     & 10000    & 10000    & 250      \\
Learning Rate Decay     & Linear   & Linear   & Linear   & Linear   & Linear   \\
Adam e                  & 1.00E-06 & 1.00E-06 & 1.00E-06 & 1.00E-06 & 1.00E-06 \\
Adam b1                 & 0.9      & 0.9      & 0.9      & 0.9      & 0.9      \\
Adam b2                 & 0.999    & 0.999    & 0.999    & 0.999    & 0.999    \\
Gradient Clipping       & 1        & 1        & 1        & 1        & 1        \\
Maximum Sequence Length & 512      & 4096     & 4096     & 8192     & 8192     \\ \bottomrule
\end{tabular}
\caption{Hyperparameters for pre-training models. Base model represents the starting model checkpoint for pre-training. We convert the NoteContrast 4k model to support longer inputs (8192) before using it to train the NoteContrast 8k model.}
\label{tab:pretraining}
\end{table*}

\begin{table*}[h!]
\centering
\begin{tabular}{@{}lllll@{}}
\toprule
Hyper-parameter         & NoteLM & NoteContrast 4k & NoteContrast 8k & NoteContrast 8k ICD \\ \midrule
Dropout                 & \multicolumn{4}{c}{0.1}                                          \\
Warmup Steps            & \multicolumn{4}{c}{200}                                          \\
Learning Rate           & \multicolumn{4}{c}{2.50E-05}                                     \\
Batch Size              & \multicolumn{4}{c}{64}                                           \\
Weight Decay            & \multicolumn{4}{c}{0.01}                                         \\
Max Steps               & \multicolumn{4}{c}{1500}                                         \\
Learning Rate Decay     & \multicolumn{4}{c}{Linear}                                       \\
Adam e                  & \multicolumn{4}{c}{1.00E-06}                                     \\
Adam b1                 & \multicolumn{4}{c}{0.9}                                          \\
Adam b2                 & \multicolumn{4}{c}{0.999}                                        \\
Gradient Clipping       & \multicolumn{4}{c}{1}                                            \\
Maximum Sequence Length & 4096   & 4096            & 8192            & 8192                \\
Training Time           & 1 hour & 1 hour          & 2 hours         & 2 hours             \\ \bottomrule
\end{tabular}
\caption{Hyperparameters for fine-tuning NoteLM and NoteContrast models on MIMIC-III-50.}
\label{tab:finetune_50}
\end{table*}

\begin{table*}[h!]
\centering
\begin{tabular}{@{}lllll@{}}
\toprule
Hyper-parameter         & NoteLM     & NoteContrast 4k & NoteContrast 8k & NoteContrast 8k ICD \\ \midrule
Dropout                 & \multicolumn{4}{c}{0.1}                                              \\
Warmup Steps            & \multicolumn{4}{c}{200}                                              \\
Learning Rate           & \multicolumn{4}{c}{2.50E-05}                                         \\
Batch Size              & \multicolumn{4}{c}{48}                                               \\
Weight Decay            & \multicolumn{4}{c}{0.1}                                              \\
Max Steps               & \multicolumn{4}{c}{500}                                              \\
Learning Rate Decay     & \multicolumn{4}{c}{Linear}                                           \\
Adam e                  & \multicolumn{4}{c}{1.00E-06}                                         \\
Adam b1                 & \multicolumn{4}{c}{0.9}                                              \\
Adam b2                 & \multicolumn{4}{c}{0.999}                                            \\
Gradient Clipping       & \multicolumn{4}{c}{1}                                                \\
Maximum Sequence Length & 4096       & 4096            & 8192            & 8192                \\ 
Training Time           & 25 minutes & 25 minutes      & 50 minutes      & 50 minutes         \\ \bottomrule
\end{tabular}
\caption{Hyperparameters for fine-tuning NoteLM and NoteContrast models on MIMIC-III-rare50.}
\label{tab:finetune_rare50}
\end{table*}

\begin{table*}[h!]
\centering
\begin{tabular}{@{}ll@{}}
\toprule
Hyper-parameter         & NoteContrast 8k \\ \midrule
Dropout                 & 0.1             \\
Warmup Steps            & 2000            \\
Learning Rate           & 5.00E-05        \\
Batch Size              & 192             \\
Weight Decay            & 0.01            \\
Max Steps               & 10000           \\
Learning Rate Decay     & Linear          \\
Adam e                  & 1.00E-06        \\
Adam b1                 & 0.9             \\
Adam b2                 & 0.999           \\
Gradient Clipping       & 1               \\
Maximum Sequence Length & 8192            \\
Training Time           & 20 hours        \\ \bottomrule
\end{tabular}
\caption{Hyperparameters for fine-tuning NoteContrast 8k ICD on MIMIC-III-full.}
\label{tab:finetune_full}
\end{table*}

\begin{table*}[h!]
\centering
\begin{tabular}{@{}llll@{}}
\toprule
Dataset                & Train         & Dev         & Test      \\ \midrule
MIMIC-III-full         & 47,723        & 1,631       & 3,372     \\
MIMIC-III-50           & 8,066         & 1,573       & 1,729     \\
MIMIC-III-rare50       & 249           & 20          & 142        \\ \bottomrule
\end{tabular}
\caption{Number of samples in each split of MIMIC-III-full, MIMIC-III-50 and MIMIC-III-rare50 datasets}
\label{tab:mimic-dataset}
\end{table*}

\end{document}